%% file: root.tex
\documentclass[letterpaper, 10 pt, conference]{ieeeconf}  %

\IEEEoverridecommandlockouts                              %

\usepackage{amssymb}

\usepackage{times}
\usepackage{epsfig}
\usepackage{graphicx}
\usepackage{amsmath}
\usepackage{amssymb}

\usepackage[dvipsnames]{xcolor}
\usepackage{multirow}
\usepackage{colortbl}
\usepackage{color}
\usepackage{hyperref}
\usepackage{xspace}
\usepackage{tabularx}

\usepackage{tikz}
\usetikzlibrary{calc,shapes,arrows,backgrounds}
\usepackage{pgfplots}
\pgfplotsset{compat=1.16}
\usepackage{pgfplotstable}
\usepackage{tikzscale}
\usepackage{csvsimple}

\usepackage{siunitx}
\usepackage{booktabs}
\usepackage{makecell}
\usepackage{eqnarray}
\usepackage{bbm}
\usepackage{textcomp}
\usepackage{gensymb}
\usepackage{pifont}
\usepackage{subcaption}
\usepackage{dsfont}
\usepackage{balance}
\usepackage{hyperref}
\usepackage[percent]{overpic}

\newcolumntype{Y}{>{\centering\arraybackslash}X}
\newcolumntype{Z}{>{\raggedleft\arraybackslash}X}

\makeatletter
\DeclareRobustCommand\onedot{\futurelet\@let@token\@onedot}
\def\@onedot{\ifx\@let@token.\else.\null\fi\xspace}
\makeatother

\input{colors.tex}

\input{macros.tex}

\title{\LARGE \bf
2D vs. 3D LiDAR-based Person  Detection on Mobile Robots
}

\author{Dan Jia \hspace{2.5em} Alexander Hermans \hspace{2.5em} Bastian Leibe%
\thanks{*All authors are with the Visual Computing Institute, RWTH Aachen {\tt\footnotesize \{jia, hermans, leibe\}@vision.rwth-aachen.de} \qquad
An earlier version with additional experiments can be found at \url{https://arxiv.org/abs/2106.11239v1}.}%
}

\begin{document}

\maketitle
\thispagestyle{empty}
\pagestyle{empty}

\input{paper.tex}

\bibliographystyle{ieee_fullname}
\bibliography{abbrev_short,mybib}
\end{document}

%% file: colors.tex
\definecolor{new_purple}{HTML}{D175F0}
\definecolor{new_gray}{HTML}{777777}
\definecolor{new_pink}{HTML}{FF69B4}
\definecolor{new_brown}{HTML}{8B4513}
\definecolor{new_blue_dark}{HTML}{1269B0}
\definecolor{new_red_dark}{HTML}{BF534F}
\definecolor{new_green_dark}{HTML}{1e942e}
\definecolor{new_yellow_dark}{HTML}{c99436}
\definecolor{new_cyan_dark}{HTML}{3ab7c7}
\definecolor{new_orange_dark}{HTML}{f05d02}

\definecolor{new_red}{HTML}{ea4335}
\definecolor{new_green}{HTML}{34a853}
\definecolor{new_blue}{HTML}{4285f4}
\definecolor{new_yellow}{HTML}{fbbc04}
\definecolor{new_orange}{HTML}{ff6d01}
\definecolor{new_lightblue}{HTML}{46bdc4}

\definecolor{gs_red}{HTML}{ea4434}
\definecolor{gs_blue}{HTML}{4285f4}
\definecolor{gs_yellow}{HTML}{fbbc04}
\definecolor{gs_green}{HTML}{34a853}
\definecolor{gs_orange}{HTML}{ff6d01}
\definecolor{gs_blue_light}{HTML}{46bdc4}

%% file: macros.tex
\newcommand{\reffig}[1]{Fig.~\ref{#1}}
\newcommand{\reftab}[1]{Table~\ref{#1}}

\newcommand*\widebar[1]{%
   \hbox{%
     \vbox{%
       \hrule height 0.5pt %
       \kern0.3ex%
       \hbox{%
         \kern-0.1em%
         \ensuremath{#1}%
         \kern-0.1em%
       }%
     }%
   }%
} 

\def\eg{\emph{e.g}\onedot}

\def\etc{\emph{etc}\onedot} \def\vs{\emph{vs}\onedot}
 
\def\etal{\emph{et al}\onedot}

%% file: paper.tex
\begin{abstract}
Person detection is a crucial task for mobile robots navigating in human-populated environments.
LiDAR sensors are promising for this task, thanks to their accurate depth measurements and large field of view.
Two types of LiDAR sensors exist: the 2D LiDAR sensors, which scan a single plane, and the 3D LiDAR sensors, which scan multiple planes, thus forming a volume.
How do they compare for the task of person detection?
To answer this, we conduct a series of experiments, using the public, large-scale JackRabbot dataset and the state-of-the-art 2D and 3D LiDAR-based person detectors (DR-SPAAM and CenterPoint respectively).
Our experiments include multiple aspects, ranging from the basic performance and speed comparison, to more detailed analysis on localization accuracy and robustness against distance and scene clutter.
The insights from these experiments highlight the strengths and weaknesses of 2D and 3D LiDAR sensors as sources for person detection, and are especially valuable for designing mobile robots that will operate in close proximity to surrounding humans (e.g. service or social robot).

\end{abstract}

\section{Introduction}
\label{sec:introduction}

Person detection is an important task in many robotic applications, including safe autonomous navigation in human-populated environments or during human-robot interactions.
LiDAR sensors are well-suited for person detection, thanks to their accurate depth measurements, long sensing range, and large field of view.
There are many successful object detection methods using 3D LiDAR sensors in driving scenarios~\cite{Zhou17CVPR,Yan18Sensors,Yang19ICCV,Lang19CVPR,Shi20PAMI,Shi20CVPR,Yang20CVPR,Yin21CVPR}, where pedestrian is typically one of the classes to be detected.
Meanwhile, in mobile robot scenarios persons have successfully been detected using range scans from 2D LiDAR sensors \cite{Arras07ICRA,Pantofaru10ROS,Leigh15ICRA,Beyer18RAL,Jia20IROS,Jia21ICRA}.

In this work, we present a comparative study on 2D and 3D LiDAR-based person detection.
Properly understanding the performance differences (detection accuracy and runtime) between these two sensor types plays an important role for well-informed robot design~\cite{Linder21IROS}.
We focus specifically on scenarios that are encountered by \eg social robots or service robots, which are becoming increasingly relevant in recent years.\footnote{These scenarios differ significantly from driving scenarios in terms of density and proximity of surrounding persons, sensor height, encountered objects, \etc.}
Using the state-of-the-art CenterPoint~\cite{Yin21CVPR} and DR-SPAAM~\cite{Jia20IROS} as representatives for methods based on 3D and 2D LiDAR sensors, we conduct a series of experiments on the large-scale, publicly available JackRabbot dataset~\cite{Martin21PAMI}, comparing the performance differences between these two sensor types.

Our experiments reveal that the 3D LiDAR-based CenterPoint provides superior detection accuracy compared to the 2D LiDAR-based DR-SPAAM, but when only considering  persons visible in the sensor data, these two methods perform on par.
In applications where detecting nearby persons (which are most likely to be visible) is the primary requirement, 2D LiDAR sensors, which are often readily available on many mobile robots for mapping and safety purposes, can be a sufficient detection source.
Additionally, 2D LiDAR-based DR-SPAAM has the advantage of higher inference speed, which is beneficial for mobile robots with limited onboard compute or power.
General purpose person detection, however, is better accomplished with 3D LiDAR sensors, which scan a 3D sensing volume (not limited to a single scan plane as 2D LiDAR sensor) and are thus more robust towards occlusion.
Our experiments also show that both methods can deliver well-localized detections and have similarly robustness against dense crowds.
A closer examination on positive and negative predictions show that a non-negligible amount of persons are only detected by one of the detectors, implying that one should opt for an ensemble-based approach when maximum reliability is the primary design objective.  

\begin{figure}[t]
    \begin{overpic}[width=1.015\linewidth]{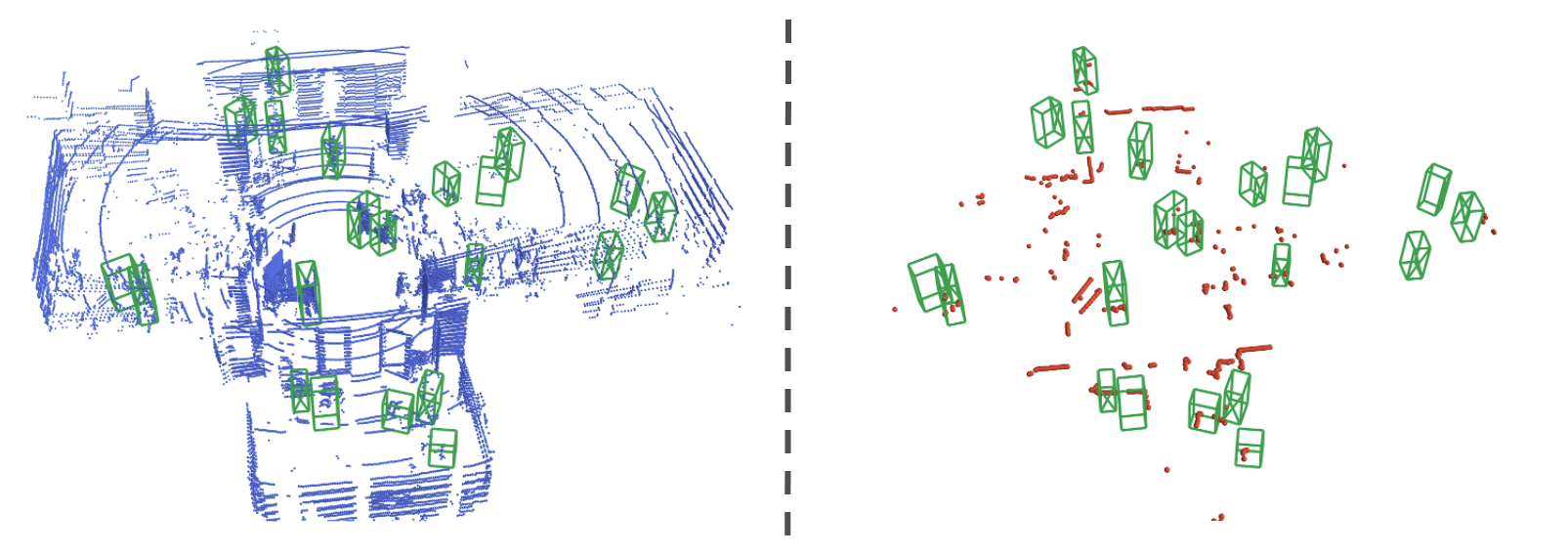}
    \put (40,5) {3D}
    \put (90,5) {2D}
    \end{overpic}
    \caption{
        Scans from a 3D and a 2D LiDAR sensor with person bounding boxes annotations.
        The vast difference between the scans are apparent, but how do these two sensors compare, when it comes to person detection?
        Understanding how these sensors affect this task is crucial for designing robots that are intended to operate around humans.
    }
    
    \label{fig:teaser}
\end{figure}

\section{Related Work}
\label{sec:related_work}

\textbf{Person detectors based on 3D LiDAR data} estimate 3D bounding boxes of persons in a scene.
Most such detectors have been designed for multi-class detection relying on autonomous driving datasets~\cite{Geiger13IJRR,Caesar20CVPR,Sun20CVPR,Kesten19Lyft}, in which the pedestrian class is one of the classes to be detected.
Existing deep-learning-based methods can largely be grouped into two categories.
Single-stage detectors~\cite{Engelcke17ICRA,Zhou17CVPR,Yan18Sensors,Lang19CVPR,Zhou19CORL,Liu19AAAI,Wang20ECCV,Chen20ECCV,Yang20CVPR,Zheng21AAAI,Yin21CVPR,Zheng21CVPR,Hu22AAAI} use backbone networks, either based on points~\cite{Qi17NIPS,Thomas19ICCV} or voxels~\cite{Zhou17CVPR,Lang19CVPR,Graham18CVPR,Choy19CVPR} to process the scene and generate bounding boxes directly from the extracted features. 
Two-stage detectors~\cite{Shi19CVPR,Qi19ICCV,Yang19ICCV,Shi20PAMI,Chen19ICCV,Shi20CVPR} additionally introduce a bounding box refinement stage, which pools features within the box proposals and generates refined predictions.
There exist, in addition, person detection methods that use hand-crafted features~\cite{Yan20AR}, which typically do not perform as well as the end-to-end learned detectors.

The KITTI dataset~\cite{Geiger13IJRR} is a forerunner in providing a standardized benchmark for testing LiDAR-based 3D object detectors, but it is now succeeded by newer and larger datasets, including the nuScenes dataset~\cite{Caesar20CVPR} and the Waymo Open Dataset~\cite{Sun20CVPR}.
However, these driving datasets differ from the mobile robot scenarios, where the robot moves in close proximity to humans (\eg the JackRabbot dataset~\cite{Martin21PAMI}).

In this work we use CenterPoint~\cite{Yin21CVPR} as the representative for 3D LiDAR-based detectors.
At the time of writing, it ranks second on the nuScenes benchmark~\cite{Caesar20CVPR} (the source-code for the higher ranking method~\cite{Hu22AAAI} is not available), and thus represents the current available state-of-the-art in the area of 3D detection.

There are, in addition, 3D methods developed primarily on datasets collected by scanning static indoor scenes with RGB-D cameras~\cite{Song15CVPR,Chang173DV,Dai17CVPR,Armeni17arXiv}.
While in theory some of these approaches could be adapted for detecting persons in LiDAR data, we do not experiment with them in this work.

\textbf{Person detectors based on 2D LiDAR data} focus on estimating centroids of persons in the scene, parametrized as $x, y$ coordinates on the LiDAR scan plane.
It has long been a relevant task in the robotics community, given it is crucial in order for a mobile agent to autonomously navigate in human-populated environments. 
Early approaches~\cite{Fod02ICRA,Scheutz04IROS,Schulz03IJRR} detect blobs with manually engineered heuristics, and track these blobs in sequential scans, leveraging motion as a clue for detection.
Later works~\cite{Arras07ICRA,Pantofaru10ROS,Leigh15ICRA} improved the detection stage, replacing the heuristics with learned classifiers (\textit{e.g.}~AdaBoost) on hand-crafted features, and still rely on motion-based filtering to obtain high quality detections~\cite{Leigh15ICRA,Pantofaru10ROS}.

Most recent developments~\cite{Beyer16RAL,Beyer18RAL,Jia20IROS,Jia21ICRA} resorted to deep-learning techniques by applying 1D CNNs to range data, and no longer require motion-based filtering.
The first of these approaches is the DROW detector~\cite{Beyer16RAL}.
It was originally designed to detect walking aids, and was later extended to also detect persons~\cite{Beyer18RAL}.
The current state-of-the-art method is the DR-SPAAM detector~\cite{Jia20IROS,Jia21ICRA}, which augments the DROW detector with a temporal aggregation paradigm, incorporating information over sequential LiDAR scans in order to improve the detection performance.

\textbf{Most similar to our work}, is the recent study of Linder~\etal~\cite{Linder21IROS} who compare a range of different 2D, 3D, and RGB-D-based person detectors.
While they compare a large number of older and more recent detectors, we limit our experiments to the two current state-of-the-art (and open-source) approaches and perform our experiments on the larger, public, and more diverse JackRabbot dataset~\cite{Martin21PAMI}.
Their findings with respect to 2D and 3D LiDAR-based person detection are in line with ours, however, we delve deeper into the differences, yielding interesting additional insights in how these two types of sensors compare for the task of person detection.

\section{Person Detectors}

The experiments in this paper are conducted with CenterPoint~\cite{Yin21CVPR} and DR-SPAAM~\cite{Jia20IROS}, state-of-the-art detectors based on 3D and 2D LiDAR sensors respectively.
While we here very briefly recap the main ideas of the two detectors, we refer the interested reader to the main publications for a detailed overview.

\textbf{CenterPoint}~\cite{Yin21CVPR} takes a voxelized 3D point cloud as input and uses either a VoxelNet~\cite{Zhou17CVPR} or PointPillars~\cite{Lang19CVPR} to extract a 2D feature map on the bird's-eye-view (BEV) plane.
From the extracted features, a center head is used to produce heatmaps, corresponding to $x, y$ locations of bounding box centers on the BEV plane.
This center head is supervised with 2D Gaussians produced by projecting bounding box centers onto the BEV plane, together with the focal loss~\cite{Lin17ICCV}.
In addition, regression heads are used to obtain $x, y$ center refinement, center elevation, box dimensions, and orientation (encoded with sine and cosine values).
These regression heads are supervised at the ground truth centers with an $L1$ loss.
In this work, we use the CenterPoint with a VoxelNet backbone for our experiments.

\textbf{DR-SPAAM}~\cite{Jia20IROS} takes as input the range component of a 2D LiDAR scan encoded using polar coordinates.
A preprocessing step is used to extract points within small angular windows (called \textit{cutouts}), and each window is normalized and passed into a 1D CNN.
The network has two branches: a classification branch, which classifies if the window center is in the proximity of a person, and a regression branch, which regresses an offset from the window center to the $x, y$ center location of the person.
The classification branch is supervised with a binary cross-entropy loss, whereas the regression branch is supervised only for positive windows, using an $L2$ loss.
Finally, the predictions from all windows are postprocessed with a distance-based non-maximum-suppression to obtain the final detections.

\section{Experimental Setup}
\label{sec:evaluation}

\subsection{Datasets}
Our main experiments are conducted with the JackRabbot Dataset and Benchmark (JRDB)~\cite{Martin21PAMI}, however, we additionally use the nuScenes dataset~\cite{Caesar20CVPR} for some pretraining experiments.

\textbf{JRDB}~\cite{Martin21PAMI} contains 54 sequences collected with a mobile robot (the \textit{JackRabbot}) moving in both indoor and outdoor environments on university campuses.
These sequences are split into 27 sequences for training and validation, and 27 for testing.
The robot is equipped with two 3D LiDAR sensors (Velodyne 16 Puck LITE), on the upper and lower part of the robot respectively, each producing scans with approximately 18,000 points.
Persons in the environment are annotated with 3D bounding boxes, with a total of roughly 1.8M annotations across the dataset.
At the moment of writing, JRDB is the only large-scale dataset that focuses on mobile robot scenarios, while also featuring LiDAR point cloud with 3D person annotations.

In addition, the JackRabbot is equipped with two 2D LiDAR sensors (SICK LMS500-20000), front and rear facing respectively.
They are mounted at the height of the lower legs, and their scans are merged to a single 360$^{\circ}$ scan, having 1091 points.
Jia \etal \cite{Jia21ICRA} used these scans to evaluate 2D LiDAR-based person detectors by synchronizing and aligning them to the 3D annotations.
In this work, we use 2D LiDAR scans from JRDB, following the same data preparation procedure.
All our reported numbers are based on the JRDB validation set obtained from the standard train-validation split.

\textbf{The nuScenes dataset}~\cite{Caesar20CVPR} contains 1,000 short sequences (approximately 20 seconds each) collected from driving vehicles.
These sequences are split into a train, validation, and test set, having 700, 150, 150 sequences respectively.
The dataset is captured using a 32 beam LiDAR, producing scans with approximately 30,000 points at 20 Hz.
Compared to JRDB, these scans cover a larger area and have significantly sparser point clouds.
Every tenth frame (0.5 seconds apart) is annotated with 3D bounding boxes, with a total of 10 classes, one of which being pedestrian.
A common practice is to combine the unlabeled scans to obtain a denser point cloud~\cite{Caesar20CVPR,Zhu19arXiv,Yang20CVPR,Yin21CVPR} during training and inference.

\subsection{Training Setup}
\label{sec:training_setup}

\textbf{For CenterPoint}, we mostly use the same hyperparameters from~\cite{Yin21CVPR}.
We follow the same training procedures and data augmentation scheme, with the AdamW optimizer \cite{Loshchilov19ICLR}, but train the network for 40 epochs with batch size 32, max learning rate $1e-3$, weight decay 0.01, and momentum 0.85 to 0.95.
For the voxelization of the input point cloud, we found that the default settings for outdoor scenarios are not optimal for JRDB.
We experiment with two settings: one where we uses the parameters for nuScenes: (0.1m, 0.1m, 0.2m) voxel grids, and limited detection range of [-51.2m, 51.2m] for the $x$ and $y$ axes, and [-5m, 3m] for the $z$ axis.
Additionally, we use a more fine-grained voxelization using a (0.05m, 0.05m, 0.2m) grid and a limited detection range of [-25.6m, 25.6m] for the $x$ and $y$ axes, and [-2m, 8m] for the $z$ axis.
This reduction in voxel size and detection range is motivated by the fact that scenes in JRDB are typically of a smaller scale compared to those in nuScenes.

When fine-tuning a network pretrained on nuScenes, we reduce the training duration to 10 epochs. This resulted in the same performance as a complete 40 epoch training schedule on top of the pretrained network. Since nuScenes provides additional time increments and reflectance features for every point, we cannot directly fine-tune checkpoints provided by the authors. Instead we retrain the network on nuScenes, using only $x, y, z$ coordinate of the points as input features, with the default training hyperparameters from~\cite{Yin21CVPR}, and fine-tune this network.

\begin{table*}[t]
    \centering
    \setlength{\tabcolsep}{3.0pt}
    \begin{tabularx}{\linewidth}{l  l c YY c YY c YY}
        \toprule
        & && \multicolumn{2}{c}{AP\textsubscript{box}} && \multicolumn{2}{c}{AP\textsubscript{BEV}} && \multicolumn{2}{c}{AP\textsubscript{centroid}}\\
        \cmidrule{4-5} \cmidrule{7-8} \cmidrule{10-11}
        & && Default & 2D-visible && Default & 2D-visible && Default & 2D-visible\\
        \cmidrule{1-11}
        \multirow{5}{*}{CenterPoint~} 
        & trained on nuScenes && 26.3 & 26.0 && 30.9 & 28.5 && 35.3 & 32.1 \\
        & ~~~+ JRDB fine-tuning && 58.2 & 64.2 && 61.2 & 67.1 && 69.5 & 75.0 \\
        & trained on JRDB && 60.0 & 68.2 && 62.6 & 69.9 && 67.9 & 75.1 \\
        & trained on JRDB (fine-grained voxelization) && 66.0 & 75.0 && 67.1 & 76.5 && 70.1 & 78.1 \\
        & ~~~+ nuScenes pretraining && 70.0 & 78.6 && 71.4 & 80.7 && 74.9 & 82.7 \\
        \midrule
        \multirow{2}{*}{DR-SPAAM~~~} & BEV && ~~34.1$^{\ast}$ & ~~58.2$^{\ast}$ && 40.8 & 67.3 && 46.6 & 74.8 \\
         & Centroid only && -- & -- && -- & -- && 47.6 & 77.2 \\
        \bottomrule 
    \end{tabularx}
    \caption{
        Performance of different CenterPoint and DR-SPAAM variants on the JRDB validation set.
        $^{\ast}$:3D Bounding boxes are obtained using predicted BEV boxes and the average box height from the training set.
        }
    \label{tab:main_comparison}
\end{table*}

\textbf{To train DR-SPAAM}, we follow the procedures from~\cite{Jia21ICRA}, and train the network for 20 epochs with a batch size of 6, using the same pre and postprocessing hyperparameters.
The original DR-SPAAM only predicts $x, y$ locations of the person, providing no information related to bounding boxes. 
In order to compare with 3D-LiDAR-based methods, we experiment with modifying the regression branch, additionally predicting the bounding box width, length, and orientation.
The width and length are parameterized as $log(w/\bar{w})$ and $log(l/\bar{l})$, where $\bar{w}=0.5$ and $\bar{l}=0.9$, which represent the average box width and length in the JRDB training set.
The orientation is parameterized by its sine and cosine as in~\cite{Yin21CVPR}.
Both the size and orientation regression are supervised with an $L$2 loss, with a weighting factor of 0.2 for the orientation.
In addition, assuming persons do not vary significantly in height, we experiment with generating 3D bounding boxes, using the predicted BEV boxes and the average height from the training set.
We did not experiment with regressing box height from the 2D LiDAR data.

\subsection{Metrics}
Following the standard for object detection, we use the Average Precision (AP) as the main evaluation metric.
We use three variants, AP\textsubscript{box}, AP\textsubscript{BEV}, and AP\textsubscript{centroid}, each having different criteria for assigning detections to ground truth boxes.
For AP\textsubscript{box}, a detection can be assigned to a ground truth, if their 3D IoU is above $0.3$ (following the JRDB convention~\cite{Martin21PAMI}).
AP\textsubscript{BEV} relaxes the requirement to a 2D IoU criterion on the bird's-eye-view boxes, discarding the requirement related to box height and elevation.
AP\textsubscript{centroid} focuses on the center localization only, assigning a detection to a ground truth if their $x, y$ location difference is smaller than $0.5$m~\cite{Leigh15ICRA,Beyer18RAL,Jia20IROS}, dropping the requirements on box size and orientation.
In all these three variants, a ground truth can only be matched with one detection.

Since ground truth boxes have different visibility to 3D and 2D LiDAR sensors, we compute all three APs under two settings.
In the first setting, we evaluate detections against ground truth boxes that contain at least 10 points from the 3D LiDAR.
This corresponds to the default evaluation procedure of JRDB and we thus refer to it as the \textit{default} evaluation.
In the second setting, which we term \textit{2D-visible}, we evaluate detections against ground truth boxes which have at least 5 points from 2D LiDAR within a $0.5$m radius from the box $x, y$ centroid.
These two evaluation settings enable us to examine the detector performance from both types of sensors, independent of factors that cause persons to be completely invisible to the sensor (thus being impossible to detect).

\section{Results}

\subsection{2D vs. 3D LiDAR-based Detection Performance}

The performance of CenterPoint and DR-SPAAM are compared in~\reftab{tab:main_comparison}.
Under the default evaluation setting, DR-SPAAM trails CenterPoint trained on JRDB (with fine-grained voxelization) by a significant 31.9\%~AP\textsubscript{box} and 26.3\%~AP\textsubscript{BEV} (the gap on AP\textsubscript{box} is greater since DR-SPAAM can not estimate the bounding box height).
However, when evaluated against ground truth that is visible to 2D LiDAR sensors, this gap shrinks to 9.2\%~AP\textsubscript{BEV}.
For the AP\textsubscript{centroid}, DR-SPAAM and CenterPoint differ by 3.3\%.
A DR-SPAAM that regresses centroid only (as it originally is in~\cite{Jia20IROS}) has a further improved performance, leaving a small performance gap of 0.9\%~AP\textsubscript{centroid} to CenterPoint trained only with JRDB.

These numbers show that, when only the centroid locations of visible persons are concerned, detectors using 2D or 3D LiDAR sensors have a similar performance, despite the vast information gap between the detector input (a simple planar scan of the legs \vs a full-body surface scan).
Thus, for tasks like person avoidance or following, a 2D LiDAR sensor is sufficient.
General purpose detection, however, is better carried out with 3D LiDAR-based detectors, since it is easy for persons to be fully occluded, and thus impossible to detect, from the scan plane of 2D LiDAR sensors.
When designing robots, the choice between 2D and 3D LiDAR sensors should be made with task requirements in view.

The first row of~\reftab{tab:main_comparison} furthermore shows that a quite big domain gap exists between autonomous driving and mobile robot scenarios.
There are many pre-trained 3D LiDAR-based detectors for driving scenarios, but it is unlikely that they will perform well on robotic tasks involving close contact with persons.
Fortunately, this domain gap can be bridged by fine-tuning and adapting suited voxel sizes.
These numbers highlight the importance of training on data similar to that will be encountered during deployment.
For 2D LiDAR sensors, estimates about this domain gap are not directly possible, since they are rarely used in outdoor driving scenarios.

\subsection{Detailed Detector Comparisons}
The remainder of this section focuses on comparison of CenterPoint and DR-SPAAM in specific aspects. The best variants of each detector (CenterPoint pretrained on nuScenes, fine-tuned on JRDB with fine-grained voxelization, and centroid-only DR-SPAAM) are used for these comparisons with AP\textsubscript{centroid} as the evaluation metric.

\begin{figure}
    \centering
    \includegraphics[width=\linewidth]{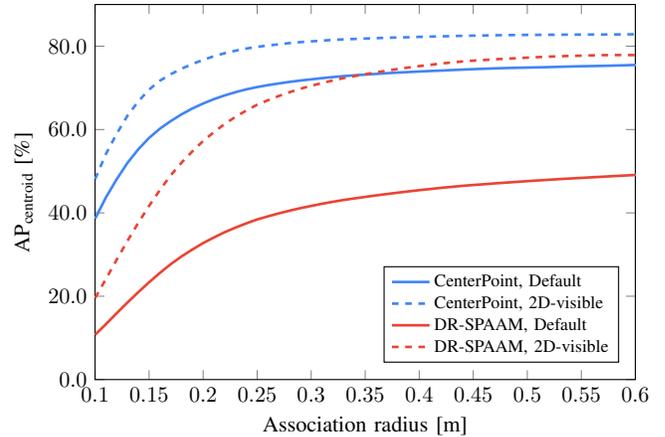}
    \caption{
        AP\textsubscript{centroid} evaluated at different ground truth association radii.
        The CenterPoint curves are slightly steeper at the start, suggesting a better localization of the detections.
    }
    \label{fig:localization_accuracy}
\end{figure}

\subsubsection{Localization Accuracy}
The default ground truth association radius of 0.5m used by AP\textsubscript{centroid} does not capture how well localized the different detections are.
To further evaluate the localization accuracy, we run the evaluation with a varying association radius and plot the development of the performance in \reffig{fig:localization_accuracy}.
Note that for this evaluation the gradient of a curve is interesting and not so much the absolute performance.

After a quick increase, the performances slowly saturate, suggesting that most detections are made with a reasonable accuracy.
The main difference between CenterPoint and DR-SPAAM is the slope of the curves for small association radii.
Here the performance for CenterPoint increases faster, suggesting a slightly better localization of its detections.
Furthermore, the CenterPoint curves are almost saturated after an association radius of $\sim$0.3m, indicating an upper bound for its localization inaccuracy.
Here the DR-SPAAM curves are still slightly increasing, meaning that some of the detections are less well localized.
Note that small inaccuracies exist in the annotations.
Thus, detection scores obtained with an overly small association radius (\eg lower than 0.1m) do not provide meaningful information.

\subsubsection{Distance-based Evaluation}

\begin{figure}
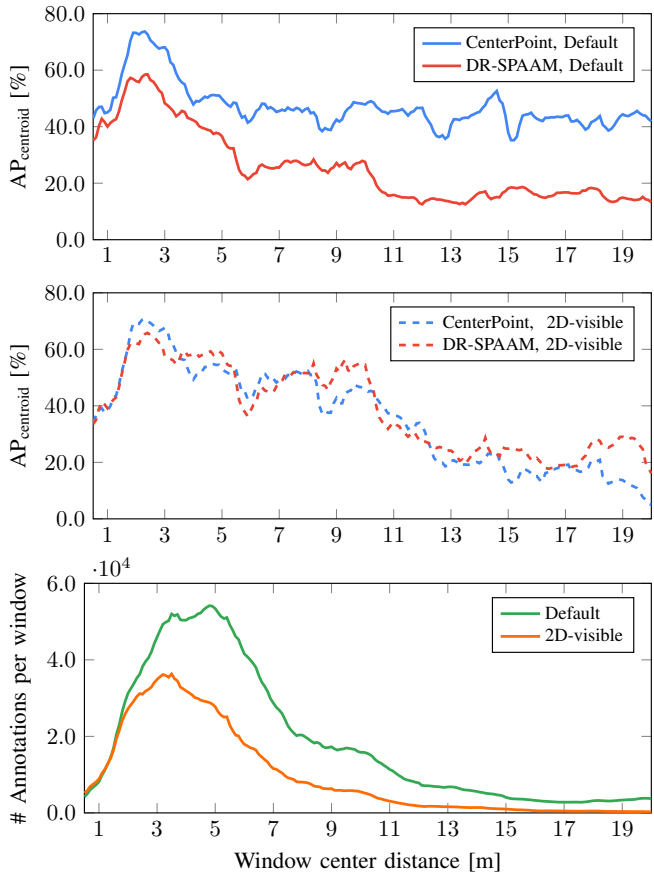

    \centering
    \includegraphics[width=\linewidth]{pics/tikz/range_full.tikz}\\[3pt]
    \includegraphics[width=\linewidth]{pics/tikz/range_filtered.tikz}\\[3pt]
    \includegraphics[width=\linewidth]{pics/tikz/range_gt_count.tikz}
    \caption{
        AP\textsubscript{centroid} evaluated within a 2m window at different distances from the sensor.
        The top and middle plot show how the performance develops across the distance for the default and 2D-visible evaluation respectively.
        The bottom plot shows the number of annotations within the windows for both evaluations.
    }
    \label{fig:performance_vs_location}
\end{figure}

The density of LiDAR data varies significantly with the distance to the sensor, which can affect the performance of detectors.
Because of that we run an additional sliding-window evaluation, where we only consider ground truth annotations within a 2m window at varying distances from the sensor.
For a meaningful evaluation we also constrain the detections to be in or near the evaluation window, as to not report unnecessary false positives.
We do this by extending the window by 0.5m (the association radius) to the front and back and only consider detections within this larger window.
This will still introduce additional false positives but also increase the true positive count, effectively being a trade-off between precision and recall.
For this reason the overall performance is lower than reported in \reftab{tab:main_comparison}, but here the relative performance between the two detectors is the interesting part.

The top and middle plots in \reffig{fig:performance_vs_location} shows how the detector performances behave for this experiment.
In all evaluations the performance is best within the first few meters and drops at higher distances.
For the default evaluation, CenterPoint performs significantly more stable across the whole range, whereas the DR-SPAAM performance decreases more at higher ranges.
This means that CenterPoint is better equipped to deal with the large variation of point densities, potentially due to the fact that objects are still visible to multiple LiDAR beams, whereas the 2D LiDAR only uses a single beam.
For the 2D-visible evaluation, the performance of both detectors is surprisingly similar across the whole range, suggesting that for visible persons, both detectors have similar range-robustness.

The bottom plot in \reffig{fig:performance_vs_location} shows how many ground truth annotations are present within the windows for both types of evaluations.
Here it becomes apparent that, from 2m and onward, a significant number of the persons are invisible to 2D LiDAR scans, and thus impossible to be detected.
The largest chunk of annotations is found within the first 12m.
After that only few annotations remain, for both the default and 2D-visible evaluation, and the evaluation becomes less meaningful.
This highlights another difference between JRDB and autonomous driving datasets, where persons are regularly perceived at much greater distances.

\subsubsection{The Influence of Crowds}

\begin{figure}
    \centering
    \includegraphics[width=\linewidth]{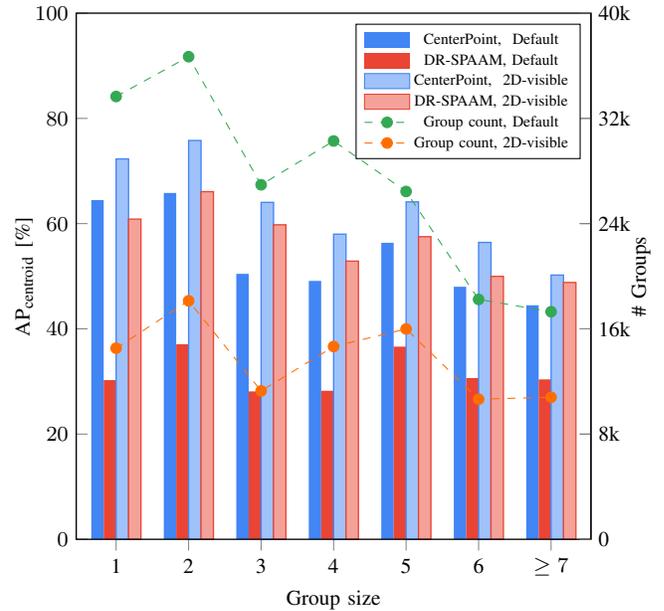}
    \caption{
        Evaluation split up by how crowded it is around annotations.
        While the performance for larger group sizes decreases a little, the effect is less pronounced than might be expected.
        We additionally plot how many groups of a specific size are present in the data, showing that JRDB is indeed fairly crowded.
    }
    \label{fig:crowd_based_evaluation}
\end{figure}

People are often perceived while interacting amongst each other, typically resulting in them being quite close together.
This results in many occlusions and could potentially lead to a more difficult task of detecting the separate persons properly.
To investigate how the detection performance changes with the object density of a scene, we look at a radius of 1.5m around every ground truth annotation and count how many other annotations can be found.
We categorize all annotations by this ``group size'' and perform a separate evaluation for all the different sizes.
\reffig{fig:crowd_based_evaluation} shows the result of this experiment.
Even though the overall detection performance seems to be reducing with larger group sizes, the detectors do not completely fail for larger group sizes and the effect seems to be very similar for both detectors.
What \reffig{fig:crowd_based_evaluation} also shows, is that a significant amount of people in the JRDB dataset are observed in the vicinity of at least one other person making it significantly more realistic than some older datasets where one would often only observe a single person in an empty scene.

\subsubsection{Runtime}
The runtime is a typical practical constraint when deploying detectors.
Especially in robotic applications, computational resources are often limited and we cannot rely on powerful desktop GPUs.
\reftab{tab:run_time} shows the runtime of the evaluated detectors on three different platforms: a powerful desktop machine, a laptop equipped with a decent GPU, and the lower-powered Jetson AGX Xavier.
Preprocessing (\eg computing \textit{cutouts}, or voxelization) and postprocessing (\eg non-maximum-suppression) steps are included in the measurement, and no batching is used for inference.
In other words, these numbers reflect the end-to-end runtime when the detector is deployed.
Both DR-SPAAM and CenterPoint achieve real-time performance on strong desktop or laptop GPUs, but not on embedded GPUs.
In general, DR-SPAAM is roughly twice as fast as CenterPoint, with the margin being smaller on the Jetson AGX.\footnote{We additionally measured the memory usage on the Jetson AGX, 5311MB for DR-SPAAM, 6318MB for CenterPoint. These numbers are likely to vary depending on specific system setups (we used L4T 32.4.4 and PyTorch 1.6).}

\begin{table}[t]
    \centering
    \begin{tabularx}{\linewidth}{lYYY}
    \toprule
         & \small \shortstack{Desktop\\(TITAN RTX)} & \small \shortstack{Laptop\\(RTX 2080)} & \small \shortstack{Jetson AGX\\Xavier}  \\
    \midrule
        CenterPoint & ~~32.4 & 19.8 & ~~6.0 \\
        DR-SPAAM & ~~59.1 & 37.1 & ~~8.9 \\
    \bottomrule
    \end{tabularx}
    \caption{Detector inference speed (frames per second) on three different platforms.}
    \label{tab:run_time}
\end{table}

\begin{table}[t]
    \centering
    \begin{tabularx}{\linewidth}{lYYl}
    \toprule
        \shortstack[l]{Subsampling\\factor} & \shortstack{AP\textsubscript{centroid}\\(Default)} & \shortstack{AP\textsubscript{centroid}\\(2D-visible)} & FPS  \\
    \midrule
        1 & 47.6 & 77.3 & ~~8.9 \\
        2 & 46.5 & 76.5 & 18.2~~(\textcolor{gs_green}{$\times$2.0}) \\
        3 & 45.2 & 75.9 & 26.6~~(\textcolor{gs_green}{$\times$3.0}) \\ 
        4 & 43.7 & 74.9 & 32.5~~(\textcolor{gs_green}{$\times$3.7}) \\ 
        5 & 42.0 & 73.3 & 36.4~~(\textcolor{gs_green}{$\times$4.1}) \\ 
    \bottomrule
    \end{tabularx}
    \caption{DR-SPAAM performance and inference speed (frames per second) on Jetson AGX Xavier with different spatial subsampling.
    }
    \label{tab:run_time_subsampling}
\end{table}

\begin{figure*}[t]
      \newcommand{\addblock}[5]{\begin{subfigure}[b]{0.49\textwidth}\centering\includegraphics[width=0.25\linewidth]{pics/qualitative/#1.png}\hspace{-4pt}\includegraphics[width=0.25\linewidth]{pics/qualitative/#2.png}\hspace{-4pt}\includegraphics[width=0.25\linewidth]{pics/qualitative/#3.png}\hspace{-4pt}\includegraphics[width=0.25\linewidth]{pics/qualitative/#4.png}\caption{#5}\end{subfigure}}%
    \addblock{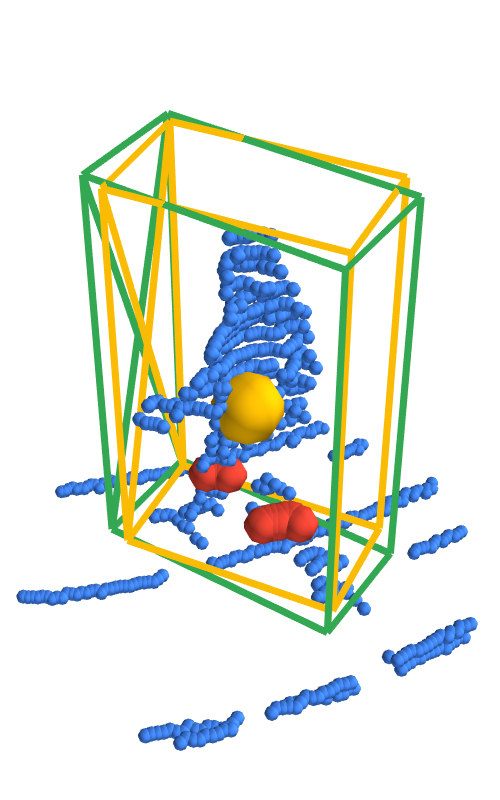}{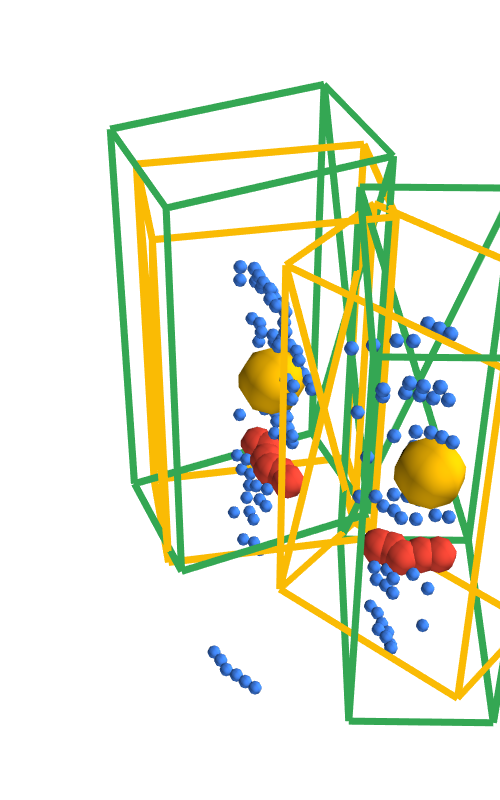}{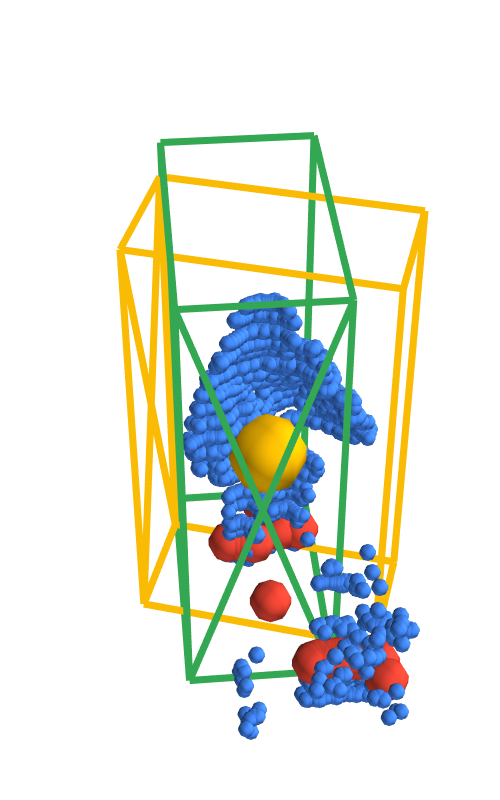}{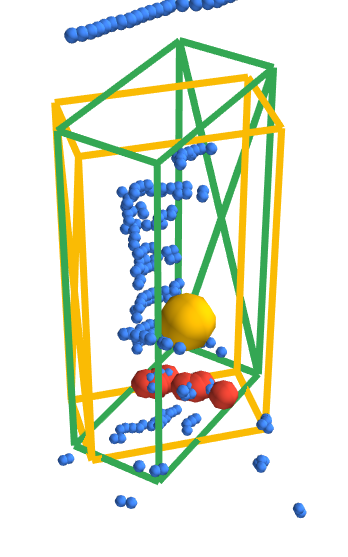}{Detected by both detectors}\hfill\addblock{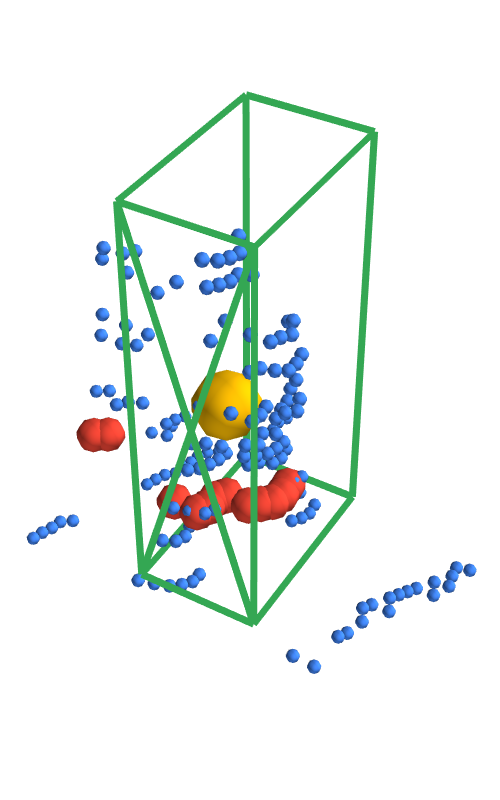}{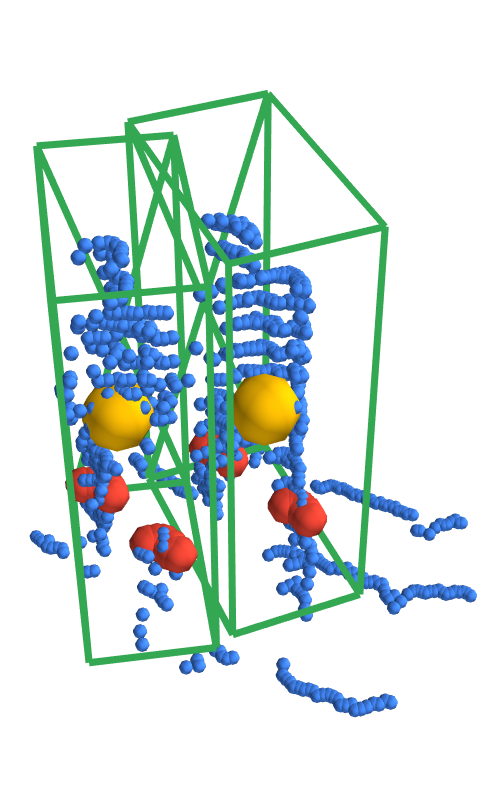}{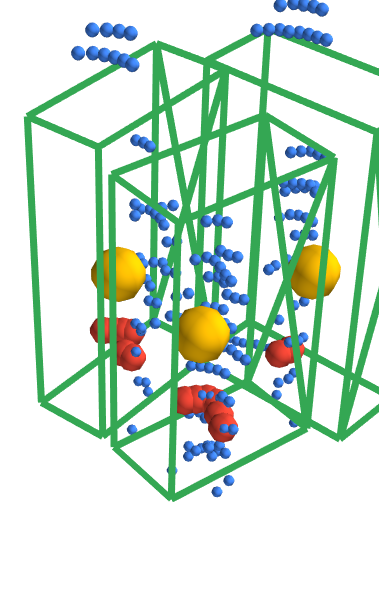}{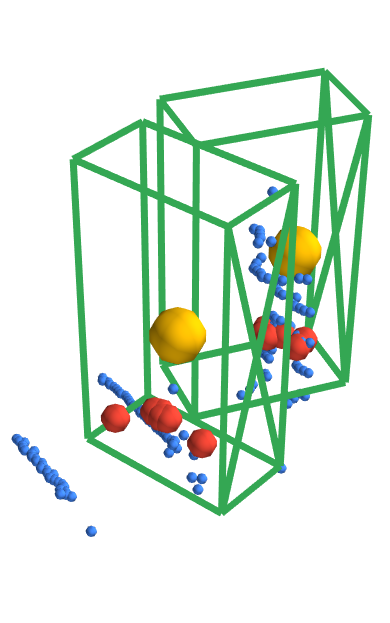}{Detected by DR-SPAAM only}\\%
    \addblock{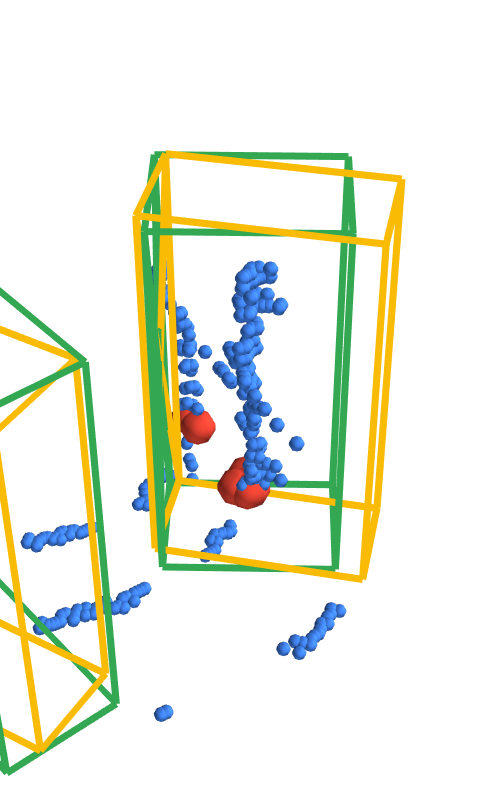}{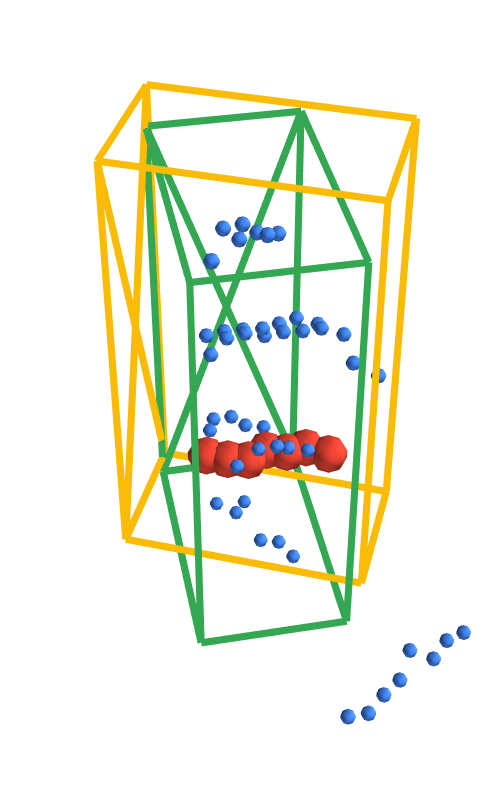}{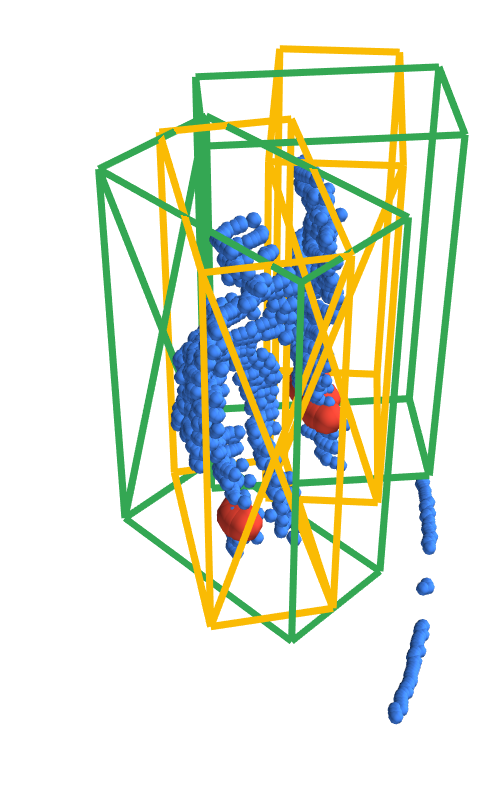}{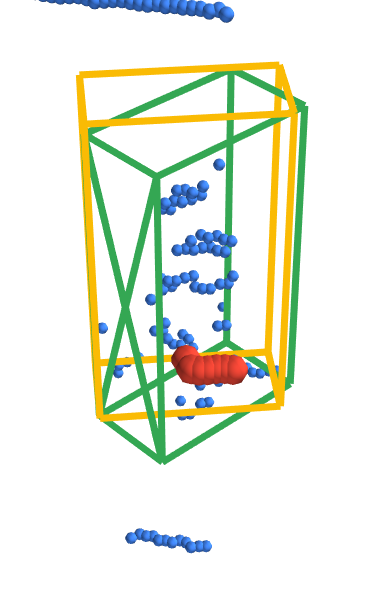}{Detected by CenterPoint only}\hfill\addblock{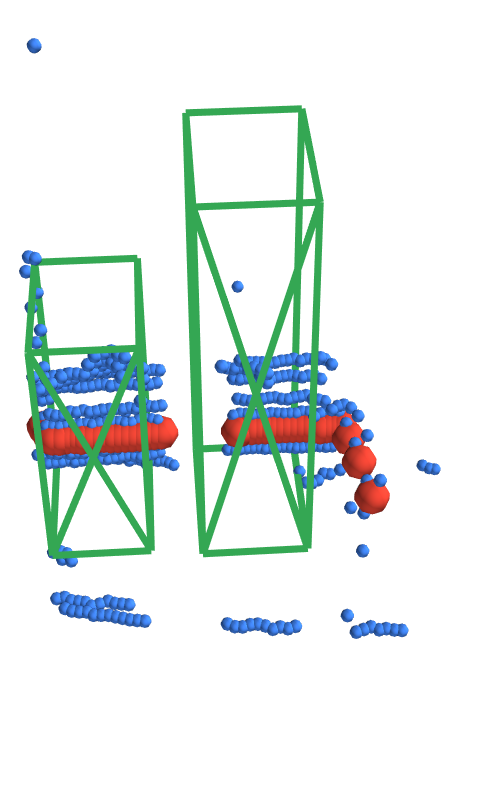}{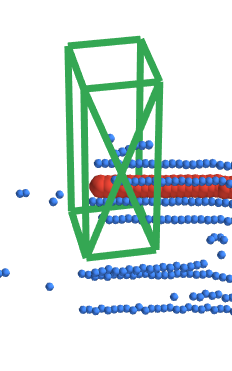}{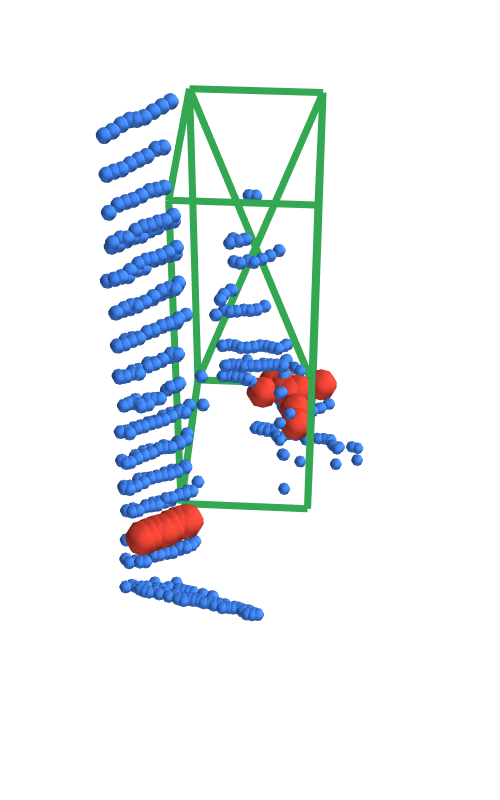}{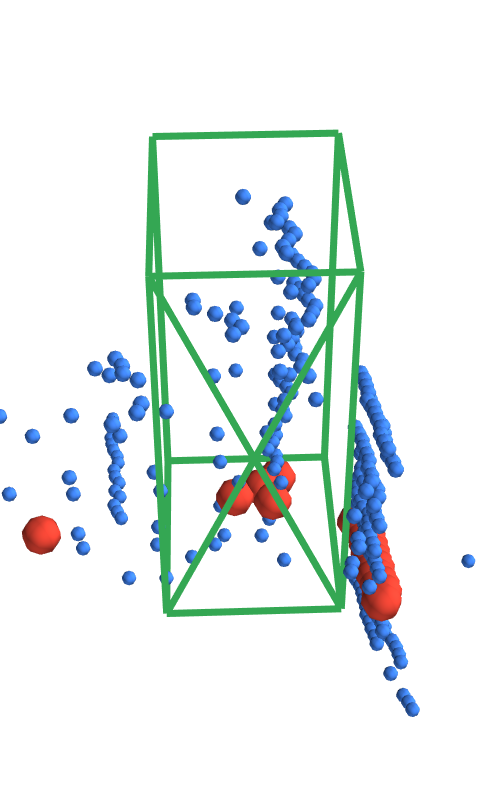}{Missed by both detectors}%
    \caption{
        Qualitative results from both detectors evaluated at equal error rates.
        Green boxes represent the ground truth, yellow boxes/spheres are detections by CenterPoint and DR-SPAAM respectively.
        The blue and red spheres are the 3D and 2D LiDAR points.
        These examples are picked from the less crowded areas to avoid visual clutter.
        Note that in many cases where one of the detectors fails, the data seems to be sufficient for a detection to be possible.
    }
    \label{fig:qualitative}
\end{figure*}

\begin{table*}[t]
    \newcommand{\tm}[2][-3pt]{\tikz[remember picture, overlay, baseline=-0.5ex]\node[xshift=0.5em, yshift=1pt, #1](#2){};}
    \newcounter{arrow}
    \setcounter{arrow}{0}
    \newcommand{\drawcurvedarrow}[3][]{%
     \refstepcounter{arrow}
     \tikz[remember picture, overlay]\draw (#2.center)edge[#1]node[coordinate,pos=0.5, name=arrow-\thearrow]{}(#3.center);
    }
    \newcommand{\annote}[3][]{%
     \tikz[remember picture, overlay]\node[#1] at (#2) {#3};
    }

    \centering
    \setlength{\tabcolsep}{2.8pt}
    \begin{tabularx}{\textwidth}{lcrYZXZXZXZX}
        \toprule
        &&&& \multicolumn{8}{c}{Detected by} \\
        \cmidrule{5-12}
        && Total GT && \multicolumn{2}{c}{Both} & \multicolumn{2}{c}{CenterPoint} & \multicolumn{2}{c}{DR-SPAAM} & \multicolumn{2}{c}{None}\\
         \midrule
         Default && 189\,579\tm[]{a} && 95\,711 & {\small(50.5\%)} & 53\,136 & {\small(28.0\%)} & 7\,212 & {\small(3.8\%)} & 33\,520 & {\small(17.7\%)}\\
         2D-visible && ~~96\,039\tm[]{b}   && 71\,619 & {\small(74.6\%)}  & 12\,432 & {\small(12.9\%)}     & 5\,263 & {\small(5.5\%)}   & ~~6\,725 & {\small~~(7.0\%)}\\
         \bottomrule
    \end{tabularx}
    \drawcurvedarrow[bend left=90, distance=3mm, -latex,line width=0.2mm]{a}{b}
    \annote[right]{arrow-1}{\small\textcolor{gs_orange}{(50.7\%)}}
    \caption{
        Detection statistics obtained using an equal error rate threshold for both CenterPoint and DR-SPAAM.
        The majority of persons is detected by both detectors, whereas a significant amount is only detected by one of the two.
    }
    \label{tab:detection_comparison}
\end{table*}

DR-SPAAM can, thanks to its \textit{cutout}-based design, obtain higher runtime by subsampling the scan, without significantly lowering the detection performance~\cite{Jia20IROS}.
\reftab{tab:run_time_subsampling} shows the performance and runtime of DR-SPAAM with different subsampling factors on the Jetson AGX.
With 3 times subsampling, DR-SPAAM can run at 26.6 FPS, while losing only 1.4\%~AP.
In applications where the onboard computation is limited, DR-SPAAM presents a more favorable trade-off between performance and runtime, compared to the more accurate, yet slower CenterPoint.

\subsection{Qualitative Results}

\reffig{fig:qualitative} shows several qualitative results of both detectors.
These results are obtained by using a detection threshold resulting in an equal error rate (EER), meaning the precision is equal to the recall for that threshold.
We show cases where both detectors, one of the two, or neither detected a person.
We specifically focus on the true positives and false negatives as these cases should contain a person.
The predicted centroids, either coming from CenterPoint or DR-SPAAM, are well-localized to the actual person.
The error in orientation estimation sometimes leads to misaligned boxes for CenterPoint, but the overlap with ground truth boxes is often sufficiently high.

Even though the numbers so far suggested that CenterPoint typically outperforms DR-SPAAM, we can here see that there are also cases where only one of the two detectors is able to detect a person.
While \reffig{fig:qualitative} shows some cases where it clearly makes sense that a detector fails, \eg due to missing points or partial occlusions.
Most of the cases where only one of the detectors fails clearly show a person though and it is unclear why it was not detected.
Interestingly, many ground truth bounding boxes missed by both detectors indeed do not look like they contain a person, potentially suggesting annotation errors.

To show these are not only a small set of cherry picked cases, \reftab{tab:detection_comparison} shows exactly how often these four different cases happen.
While CenterPoint is more often the only detector to detect a person, there is a small but significant number of people only detected by DR-SPAAM, too.
This suggests that an ensemble of both methods could be an interesting approach when both sensors are available.

\section{Discussion}
We performed all experiments with two state-of-the-art LiDAR-based detectors, while we could have tried to adapt CenterPoint to run on 2D LiDAR data, allowing for a more direct comparison.
However, this would require a significant amount of tuning to make sure CenterPoint yields a representative performance on 2D LiDAR and our goal is not to develop a new 2D LiDAR-based detector.
Instead, we here rely on an existing and well-tuned state-of-the-art method.
With our detector evaluations, we have set a lower-bound for the person detection performance on JRDB using 2D and 3D LiDAR sensors, which will be further improved by future developments.

Apart from LiDAR-based person detection, one could consider other sensor modalities.
In particular, image-based person detectors have a long history.
While the person class is now typically seen as one of many classes by most deep-learning-based detectors~\cite{Ren15NIPS,Carion20ECCV}, a whole range of person-specific detectors existed before~\cite{dalal05CVPR,Benenson12CVPR}.
Furthermore, RGB-D based person detectors are frequently used in robotics, which have the additional advantage that a person's position can be estimated \cite{Jafari14ICRA,Lewandowski19ICRA,Linder20ICRA}.
While both types of image-based detectors can perform robust person detection, they have the drawback that the field-of-view of most cameras is fairly limited, as also shown in \cite{Linder21IROS}.
Image-based detectors will likely remain a viable source for person detections, however, a single LiDAR sensor can be sufficient to cover the complete surrounding, which would typically require multiple cameras, significantly increasing compute requirements.

\section{Conclusion}

In this paper we investigated differences between 2D and 3D LiDAR-based person detection for mobile robots.
For this we perform direct comparisons betweeen the state-of-the-art CenterPoint~\cite{Yin21CVPR} and DR-SPAAM~\cite{Jia20IROS} detectors on the JackRabbot dataset~\cite{Martin21PAMI}.
We found that, when only visible persons are considered, detectors from both 2D and 3D LiDAR sensors perform on a similar level, but in 2D LiDAR sensors persons are more prone to being invisible to the sensor (and thus impossible to  detect).
While 3D LiDAR sensors overall provide a more robust solution to person detection, 2D LiDAR sensors provide better a trade-off between performance and runtime, a favorable trait for mobile robots with limited onboard computations.

Further analysis of the detectors showed that overall both their detections are well-localized and they are not significantly affected by crowds.
Since a significant number of people are only detected by one of the two detectors, if computational resources allow it, an ensemble of the detectors could be an interesting possibility to further boost the detection performance.
While one might have assumed the superiority of 3D LiDAR-based person detection, the similar performance and robustness on visible persons for these two sensor types comes as somewhat of a surprise.
We believe these insights are valuable during sensor and detector selection for robots that will navigate around persons.

\textbf{Acknowledgements:}
We would like to thank Timm Linder and Jen Jen Chung for valuable feedback.
This project was funded by the EU H2020 project ``CROWDBOT'' (779942). 
Most experiments were performed on the RWTH Aachen University CLAIX 2018 GPU Cluster (rwth0485).